%% file: example.tex
\documentclass[letterpaper, 10 pt, conference]{ieeeconf}
\IEEEoverridecommandlockouts                 
\usepackage{booktabs,siunitx}
\usepackage{algorithm}
\usepackage{algpseudocode}
\usepackage{graphicx}
\usepackage{caption}
\usepackage{amssymb}
\usepackage{wrapfig}
\usepackage{subcaption}
\usepackage{multicol}
\usepackage{balance}
\usepackage{multirow}
\usepackage[table]{xcolor}
\usepackage{threeparttable}
\usepackage{float}
\usepackage{siunitx}
\usepackage{etoolbox}
\usepackage{setspace}
\usepackage{mathtools}
\usepackage[table]{xcolor}
\usepackage{xcolor}
\usepackage[style=ieee]{biblatex}
\title{\LARGE \bf Feature Extractor or Decision Maker: Rethinking the Role of Visual Encoders in Visuomotor Policies}

\author{
    Ruiyu Wang, Zheyu Zhuang, Shutong Jin, Nils Ingelhag, Danica Kragic and Florian T. Pokorny
    \thanks{This work was partially supported by the Wallenberg AI, Autonomous Systems and Software Program funded by the Knut and Alice Wallenberg Foundation. The computations were enabled by the supercomputing resource Berzelius provided by the National Supercomputer Centre at Linköping University and the Knut and Alice Wallenberg Foundation, Sweden.}
    \thanks{All authors are from the Division of Robotics, Perception and Learning, KTH Royal Institute of Technology, Sweden.
    {\tt\small \{ruiyuw, zheyuzh, shutong, ingelhag, dani, fpokorny\}@kth.se}}
}
\begin{document}
\maketitle
\begin{abstract}
\input{body/abstract}
\end{abstract}

\section{Introduction}
    \input{body/intro}
\section{Related Work}
    \input{body/related_works}

\section{Methodology}
    \input{body/methods}
    
\section{Experiment}
    \input{body/exp}

    
\section{Conclusion and Limitation}
    \input{body/conclusion}

\balance
\printbibliography
\end{document}

%% file: body/abstract.tex
An end-to-end (E2E) visuomotor policy is typically treated as a unified whole, but recent approaches using out-of-domain (OOD) data to pretrain the visual encoder have cleanly separated the visual encoder from the network, with the remainder referred to as the policy. 
We propose Visual Alignment Testing, an experimental framework designed to evaluate the validity of this functional separation. 
Our results indicate that in E2E-trained models, visual encoders actively contribute to decision-making resulting from motor data supervision, contradicting the assumed functional separation. 
In contrast, OOD-pretrained models, where encoders lack this capability, experience an average performance drop of 42\% in our benchmark results, compared to the state-of-the-art performance achieved by E2E policies. 
We believe this initial exploration of visual encoders' role can provide a first step towards guiding future pretraining methods to address their decision-making ability, such as developing task-conditioned or context-aware encoders.

%% file: body/intro.tex
In vision-based robotic manipulation, end-to-end (E2E) visuomotor policies which map pixel inputs directly to control signals are gaining popularity due to their ability to model complex robot behaviors and motion modalities~\cite{levine2016end, pmlrv87kalashnikov18a, pmlrv155zeng21a, pmlrv205shridhar23a}.
However, training policies from scratch is still restricted to simulation environments or requires large human demonstration datasets which are difficult to collect and scale~\cite{robomimic2021, mandlekar2023mimicgen}.

In computer vision (CV), the limited availability of training data is addressed by pre-training a universal visual encoder on large image datasets and adopting it for diverse downstream tasks~\cite{Mahajan2018ECCV, rich2014cvpr}.
Recent advances in robotic manipulation have mirrored this trend~\cite{nair2022rm, Xiao2022, ma2022vip, parisi2022the, radosavovic2022realworld}.
Specifically, they pretrain a visual encoder using out-of-domain (OOD) natural image and video datasets, such as ImageNet~\cite{imagenetcvpr09} and Ego4D~\cite{grauman2022ego4d}, capturing semantic information~\cite{ma2022vip} or temporal dynamics~\cite{nair2022rm} in images or videos. 
Subsequently, the trained encoders are used frozen (weight updates are prohibited) to extract visual features when training policies across various manipulation tasks.
Performance improvements over E2E training are reported across diverse policy and task domains, particularly in low-data regimes (5 to 100 demonstrations).

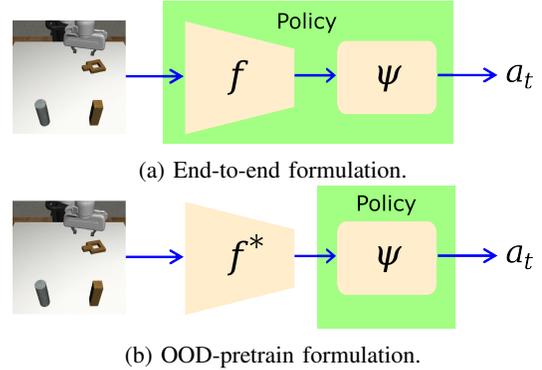
\begin{figure}
    \centering
    \input{pics/e2e_ood}
    \caption{\textbf{Formulation of Visuomotor Policies.} 
     In E2E formulation, although a visuomotor policy can be viewed as a visual encoder $f$ followed by a regression head $\psi$, the two components as a whole are treated as the policy. In OOD pretraining, a functional separation of the visual encoder emerges and only the regression head acts as a policy.}
    \label{fig:visuomotor_policy}
    \vspace{-0.3cm}
\end{figure}

Despite the growing interest and significant advancements in pretraining methods, a crucial yet persistently overlooked difference between E2E and pretraining is that OOD pretraining shifts the formulation of visuomotor policies from a unified network to a modular design, comprising a frozen universal visual encoder and a trainable regression head, as shown in \textit{Fig.}~\ref{fig:visuomotor_policy}.
This \textit{functional separation} implicitly assumes that visual encoders act only as general feature extractors, without contributing to the decision-making process, given their lack of task data supervision.

However, the validity of this functional separation and the potential benefits of pretraining is compromised if its performance margin cannot be maintained under certain circumstances.
In this paper, we establish a performance benchmark comparing models with OOD-pretrained encoders with the state-of-the-art performance achieved by E2E policies~\cite{NIPS1988812b4ba2, robomimic2021, chi2023diffusionpolicy}.
Our benchmarking results (see \textit{Tab.}~\ref{tab:benchmark}) reveal that policies utilizing frozen pretrained encoders lose their performance advantage, with an average success rate gap of 42\% in simulation.

To further investigate the causes of the observed gap and inspired by the difference in policy formulation between the two paradigms, we hypothesize that \textit{the performance gap arises because E2E-trained visual encoders play an active role in decision-making}, a capability OOD-pretrained encoders do not develop.
To verify this hypothesis, we introduce Visual Alignment Testing (VAT), an experimental framework designed to quantitatively assess the impact of visual encoders on decision-making (detailed in Section III.C). 
Additionally, we employ \textit{Fullgrad} saliency maps~\cite{Srinivas2019FullGradientRF} to qualitatively demonstrate how visual encoders contribute to decision-making through their pixel-wise activations on the visual inputs.
Experiments were conducted in both simulations and the real world.

Although the hypothesis may seem straightforward, it has to the best of our knowledge not previously been explicitly identified and verified.
Additionally, its validity is becoming increasingly important for the community as OOD pretraining is built upon the assumption of the functional separation, and reports of pretraining underperformance are growing~\cite{liu2023libero, hansen2022on, sharmalossless, lin2023spawnnet}.
This work provides initial quantitative and qualitative evidence that may in part explain the underperformance of OOD pretraining for robotic manipulation, underscoring the overlooked role of visual encoders in visuomotor policies. The main contributions are:
\begin{itemize}
    \item We present a performance benchmark between OOD pretrained visual encoders and the state-of-the-art performance achieved by E2E policies, showcasing a performance gap.
    \item We identify the performance gap stemming from E2E visual encoders actively contributing to the decision-making process and provide both quantitative and qualitative evidence via the proposed Visual Alignment Testing and \textit{Fullgrad} saliency map.
    \item Our results prompt a reflection on the role of visual encoders in visuomotor policies and the validity of their functional separation in pretraining that may be relevant for developing future methods in this area.
\end{itemize}

%% file: pics/e2e_ood.tex
\minipage{0.7\linewidth}
    \includegraphics[width=\linewidth]{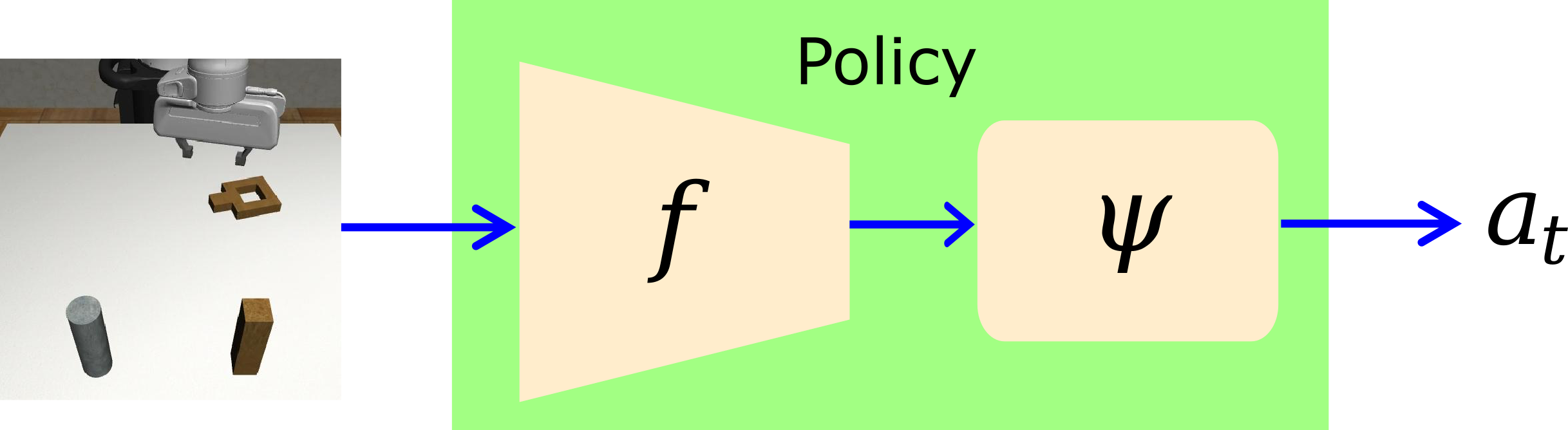}
    \subcaption{End-to-end formulation.}
    \label{fig:end}
\endminipage\hfill
\minipage{0.7\linewidth}%
    \includegraphics[width=\linewidth]{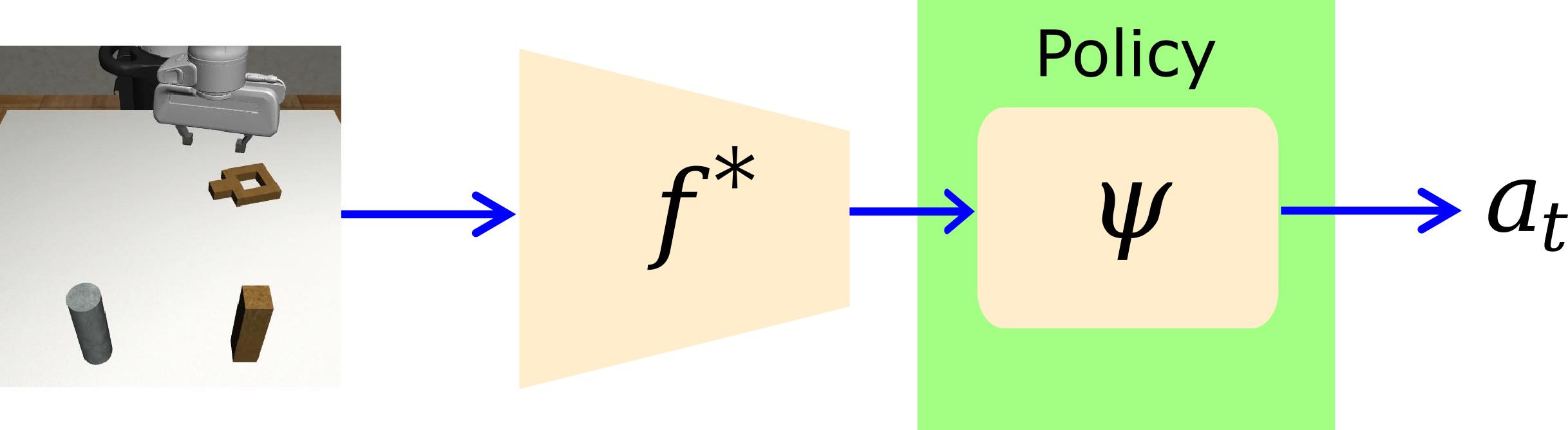}
    \subcaption{OOD-pretrain formulation.}
    \label{fig:ood}
\endminipage

%% file: body/related_works.tex
Initial attempts in applying OOD pretrained vision models to robotics directly integrate models from computer vision into the control pipeline. 
\textit{Lin et al.}~\cite{Lin2020LearningTS} adopt ImageNet~\cite{imagenetcvpr09} and COCO~\cite{cocodataset} pretrained models to predict the affordance maps for pre-defined motion primitives. 
\textit{Khandelwal et al.}~\cite{khandelwal2022} study embodied navigation-heavy tasks using frozen visual encoders from CLIP~\cite{Radford2021LearningTV}. 
\textit{Zhu et al.}~\cite{zhu2022viola} obtain general object proposals from a pretrained visual model. 
These works incorporate pretrained visual encoders into their modular-designed frameworks, using them as low-level visual features enriched with semantic priors.

Recent works have explored developing a universal encoder for manipulation which is trained on OOD datasets and leverages the frozen encoder for policy regression~\cite{radosavovic2023robot, Yao2022MaskedCR, ChaneSane2023LearningVP, Wu2023PolicyPF, zeng2024learning, karamcheti2023voltron}.  
PVR~\cite{parisi2022the} find that encoders pretrained on computer vision tasks can be transferred to behavior cloning. 
MVP~\cite{Xiao2022}, R3M~\cite{nair2022rm} and VIP~\cite{ma2022vip} investigate self-supervised representation learning for control. 
Specifically, MVP trains a masked autoencoder (MAE)~\cite{He2021MaskedAA} on natural images as the visual encoder. 
\textit{Radosavovic et al.}~\cite{radosavovic2022realworld} extend this idea from simulation to real robot imitation learning. 
R3M uses a combination of time-contrastive learning and video-language alignment to capture environmental dynamics and semantics. 
VIP generates representations and dense visual rewards by reforming it into a value function optimization problem.

Several efforts have been made to benchmark the advances in OOD-pretrained encoders for manipulation. 
\textit{Jing et al.}~\cite{jing2022exploring} explore three components in pretraining: datasets, models and training objectives.  
A more detailed dataset-centric analysis for pre-training is proposed by \cite{dasari2023an}. 
\textit{Hu et al.}~\cite{Yingdong2023for} conduct empirical evaluations for various pre-trained vision models under different policies.
However, these benchmarking studies primarily focus on comparing the performance of different pretrained encoders, without considering E2E-trained baselines.
\textit{Hansen et al.}~\cite{hansen2022on} show a simple Learning-from-Scratch baseline is competitive with state-of-the-art pretrained models. 
However, the authors attribute the performance gap between pretrained and E2E models to domain gaps, without a deeper investigation into the underlying causes.
Additionally, the role of visual encoders in visuomotor policies remains underexplored.
This paper takes steps towards addressing these gaps by benchmarking pretrained encoders against their E2E counterparts and providing quantitative and qualitative evidence into the role of visual encoders.

%% file: body/methods.tex
\begin{table}
    \vspace{0.3cm}
    \input{tables/pretrain_overview}
    \caption{\textbf{Overview of Pretrained Encoders.} We summarize the pretrained encoders considered in the benchmark comparison in terms of training objectives, model architectures, numbers of parameters (in millions), and training datasets. MVP uses a combination of the Ego4D, ImageNet and HoI~\cite{Xiao2022} datasets for the training.}
    \label{tab:overview_pretrain}
    \vspace{-0.3cm}
\end{table}
\subsection{Preliminary}
Learning-based robotic manipulation commonly develops visuomotor policies using reinforcement learning (RL)~\cite{kober2013reinforcement} or imitation learning (IL)~\cite{hussein2017imitation}. 
In this paper, we primarily focus on IL. 
An IL policy learns action distribution from expert demonstration data $(\mathcal{O}, \mathcal{P}, \mathcal{A})$, including image observation $\mathcal{O}$, proprioceptive state $\mathcal{P}$ and robot action $\mathcal{A}$. 
Under the E2E-training scenario, given an observation state $o_{t}\in\mathcal{O}$ and a proprioceptive state $p_{t}\in\mathcal{P}$, a policy $\pi$ directly maps the states to robot actions $\hat{a}_{t} = \pi (o_{t},p_{t})$. 
The policy network $\pi$ can be viewed as two parts: a visual encoder $f_{\theta}$ which extracts visual features from $o_{t}$ and a regression head $\psi_{\phi}$ which predicts the actions, $\hat{a}_{t}=\psi_{\phi}(f_{\theta}(o_{t}), p_{t})$. 
The weights $\theta\in\mathbb{R}^{n_{1}}$ and $\phi\in\mathbb{R}^{n_{2}}$ are updated simultaneously to minimize the objective $||a_{t} - \hat{a}_{t}||_{i},\;\text{for}\; i=1,2$. 
For OOD pretraining, a pretrained visual encoder $f^{*}$ is applied frozen to image inputs and only $\phi$ is updated during training. 

\subsection{Benchmark Setting}
The performance of visuomotor policies is highly sensitive to task complexity and policy capacity~\cite{robomimic2021, mandlekar2023mimicgen}. 
To better anchor the performance of pretrained encoders, we establish a benchmark where end-to-end policies achieve state-of-the-art performance~\cite{chi2023diffusionpolicy}.
Below, we outline the three components of our primary focus: pretrained encoders, end-to-end baselines and testing environments. 

\subsubsection{Pretrained Encoders}
We evaluate the performance of three state-of-the-art pretrained encoders designed for control: R3M~\cite{nair2022rm}, MVP~\cite{radosavovic2022realworld}, and VIP~\cite{ma2022vip}. 
Additionally, we include an ImageNet-pretrained ResNet18~\cite{he2015deep} due to its widespread use in CV. 
Other popular pretrained CV models are excluded, as their performance has already been extensively benchmarked in prior studies~\cite{Yingdong2023for}.
During the policy training phase, all encoders are frozen, adhering to the original evaluation protocols without modifications. 
For models with multiple variants, we adopt the smallest publicly available version to ensure a fair comparison.
The pretrained encoders are evaluated in combination with various regression heads, as detailed in Section III.B.2. 

\begin{figure*}[th]
    \centering
    \vspace{0.3cm}
    \scalebox{0.95}{
    \includegraphics[width=1.0\textwidth]{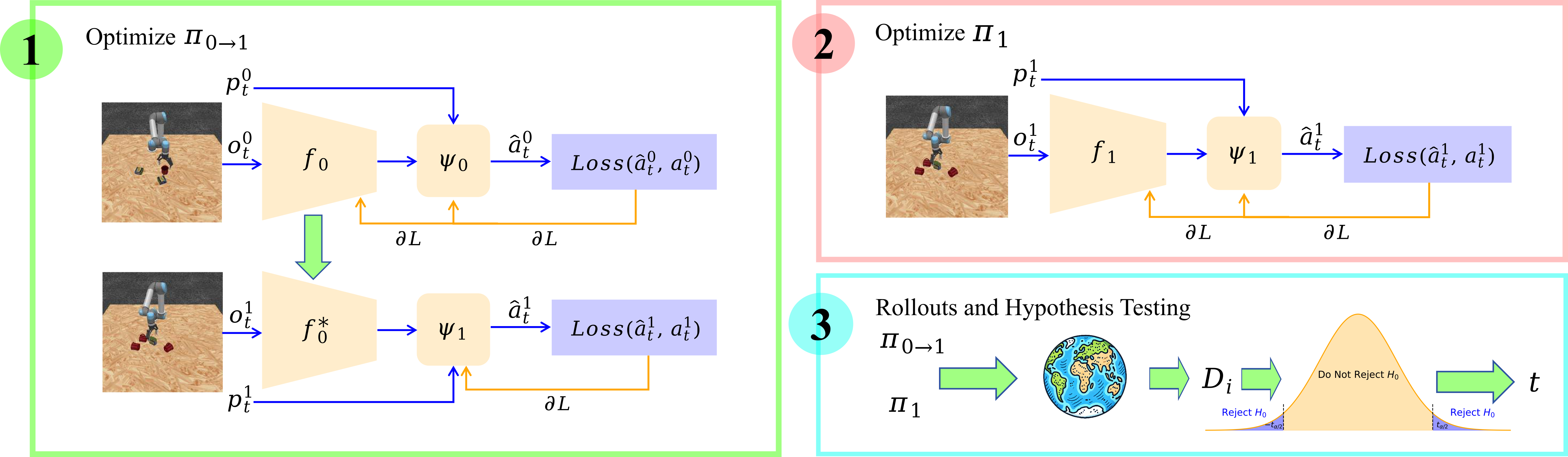}
    }
    \caption{\textbf{Visual Alignment Testing.} 
    The experimental framework we propose provides quantitative evidence that E2E-trained visual encoders play an active role in decision-making, detailed in Section III.C.}
    \vspace{-0.3cm}
    \label{fig:vat_pipeline}
\end{figure*}

\begin{table}
    \input{tables/task_overview}
    \caption{\textbf{Task Summary.} ActD: robot action dimension, ObsD: image observation dimension, Data: number of demonstrations, Step: max rollout steps in simulation during evaluation, RichC: whether the task requires contact-rich manipulation, LongH: whether the task requires learning multiple behaviors together. Tasks for benchmarking are marked in blue, while the others are used in VAT, see \textit{Fig.}~\ref{fig:vat_task}.}
    \vspace{-0.3cm}
    \label{tab:overview_task}
\end{table}

\subsubsection{End-to-end Baselines}
All E2E baselines utilize a visual encoder with the same structure, followed by different regression heads.
Existing works on pretraining predominantly evaluate the pretrained encoders in combination with a simple behavior cloning (BC) algorithm. 
However, prior studies indicate a significant performance gap between BC and other history-dependent models~\cite{robomimic2021, mandlekar2023mimicgen}, and BC often fails to model multi-modality compared to implicit policies~\cite{florence2021implicit}. 
To underscore the impact of model capacity, we include the following algorithms as the regression head in our evaluation:
\begin{itemize}
    \item BC: An explicit model consisting of a two-layer fully connected MLP.
    \item BC-RNN~\cite{robomimic2021}: An adaptation of BC incorporates RNN that anticipates outcomes over a horizon of 10 steps.
    \item DP-C~\cite{chi2023diffusionpolicy}: A convolutional-based Diffusion Policy, an implicit model that learns the gradient of the action score function.
    \item DP-T~\cite{chi2023diffusionpolicy}: A transformer-based Diffusion Policy.
\end{itemize}

The encoder design follows that of Diffusion Policy~\cite{chi2023diffusionpolicy}. 
Specifically, a standard ResNet18 model is randomly initialized and modified by replacing average pooling with spatial softmax~\cite{robomimic2021} and substituting batch normalization with group normalization~\cite{Wu2018ECCV}. 
In tasks involving multiple camera views, each view is processed by a separate learnable visual encoder.

\subsubsection{Testing Environments} 
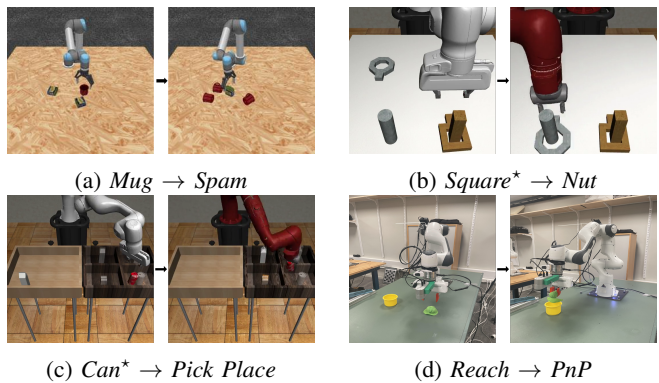
\begin{figure}[H]
    \centering
    \small
    \input{pics/transfer_task}
    \caption{\textbf{Tasks for Visual Alignment Testing.} We conduct VAT on three groups of tasks in simulation and one in the real world, illustrated in the order of $\mathcal{T}_{0}\rightarrow\mathcal{T}_{1}$, see Sec. III.C.}
    \label{fig:vat_task}
    \vspace{-0.3cm}
\end{figure}

\textit{Tab.}~\ref{tab:overview_task} provides an overview of the tasks used for benchmarking (highlighted in blue) and VAT (in black). These tasks are derived from three simulation environments and one real-world setting. An illustration of tasks used in VAT is shown in \textit{Fig.}~\ref{fig:vat_task}.

\begin{itemize}
    \item \textit{Robomimic}~\cite{robomimic2021} and \textit{MimicGen}~\cite{mandlekar2023mimicgen}: Two prominent demonstration datasets and benchmarks for robotic manipulation. We adopt \textit{Can} and \textit{Square} from Robomimic and \textit{Stack}, \textit{Pick Place} and \textit{Nut Assembly} from MimicGen. $\textit{Can}^{*}$ and $\textit{Square}^{*}$ refer to two customized datasets that are collected by replaying the proficient-human dataset of \textit{Square} and \textit{Can} in \textit{Robomimic} while adding a round nut or one random object appears in \textit{Pick Place} in each demonstration. 
    \item \textit{PushT}: A task introduced by IBC~\cite{florence2021implicit} featuring long-horizon and contact-rich dynamics. 
    \item \textit{UR5 Reach}: Our customized simulation environment where a UR5 robot reaches for the closest \textit{Mug} or a box of \textit{Spam} on a table. The dataset consists of 1000 machine-generated trajectories for each task.
    \item \textit{Franka Pick}: Real-world experiments on a Franka Emika Panda robot. In \textit{Real Reach}, the robot reaches for a deformable ball, and in \textit{Real PnP}, it picks and places the ball into a bucket. The data for these tasks were collected by a proficient human operator teleoperating the robot via a VR controller~\cite{welle2024quest2ros}.
\end{itemize}

\subsection{Visual Alignment Testing}
We hypothesise that \textit{E2E-trained visual encoders are playing an active role in decision-making}.
This hypothesis, if shown to hold, may explain the underperformance of OOD pretraining which lacks task data supervision due to the inherent functional separation.
However, the implicit nature of neural networks makes direct validation challenging.
Therefore, we propose Visual Alignment Testing, a proof-by-contradiction approach outlined below.

In imitation learning, the definition of tasks is embedded in their demonstration data $(\mathcal{O}, \mathcal{P}, \mathcal{A})$.
In most cases, two different tasks $\mathcal{T}_{0} (\mathcal{O}_{0}, \mathcal{P}_{0}, \mathcal{A}_{0})$ and $\mathcal{T}_{1} (\mathcal{O}_{1}, \mathcal{P}_{1}, \mathcal{A}_{1})$, such as picking up a cup and cutting an apple, have both distinct visual inputs and motor signals.
One special case is when two tasks \textit{share the same visual distribution\footnote{The same visual distribution refers to the distribution of visual contents.} but have different motor signals}. 
For example in \textit{Fig.}\ref{fig:task_vat_1}, both tasks—reaching for a mug and a box of spam—include the same visual contents in the scene, but the control signals direct them to interact with different objects.

Consider two tasks $\mathcal{T}_{0} (\mathcal{O}, \mathcal{P}_{0}, \mathcal{A}_{0})$ and $\mathcal{T}_{1} (\mathcal{O}, \mathcal{P}_{1}, \mathcal{A}_{1})$ that share the same visual distribution but differ in motor signals.
Two visuomotor policies $\pi_{0}$ and $\pi_{1}$ are trained end-to-end for each task respectively with successful convergence. 
Assuming that visual encoders do not play an active role in decision-making, then the encoders $f_{0}$ and $f_{1}$ should yield similar features given the same inputs, as they are trained on visual data with the same distribution.

\begin{equation}
    f_{0}(o_{t}) \sim f_{1}(o_{t}),\quad f_{0}\in\pi_{0},\;f_{1}\in\pi_{1},
    \label{eq:1}
\end{equation}
where $o_{t}$ is a image observation of $\mathcal{T}_{0}$ or $\mathcal{T}_{1}$ at time $t$.

In other words, $f_{0}$ and $f_{1}$ should be interchangeable between the two policies. Consequently, if we extract $f_{0}$ trained on $\mathcal{T}_{0}$ and apply it frozen $f_{0}^{*}$ to train another policy for $\mathcal{T}_{1}$, resulting in $\pi_{0\rightarrow 1}\coloneqq\pi_{1}|_{f=f_{0}^{*}}$ , no significant performance discrepancy between $\pi_{1}$ and $\pi_{0\rightarrow 1}$ should be observed in expectation.
\begin{equation}
    \mathbb{E}[d(\pi_{1})] \approx \mathbb{E}[d(\pi_{0\rightarrow 1})]
    \label{eq:2}
\end{equation}
where $d$ is the metric for policy performance, and $\mathbb{E}$ stands for the expectation under testing rollouts.

Therefore, for a given performance discrepancy threshold $\delta > 0$, if there exists $\epsilon >0$, such that
\begin{equation}
    \mathbb{E}[|d(\pi_{1}) - d(\pi_{0\rightarrow 1})|] > \delta + \epsilon
\end{equation}
A contradiction occurs. This would imply that the assumption of visual encoders not playing an active role in decision-making is shown to be incorrect.
A threshold $\delta$ is introduced to capture training randomness.

This chain of logic can be empirically verified through hypothesis testing applied to the collected experimental data. 
We refer to this experimental framework as Visual Alignment Testing (VAT).
Let $X_{i}$ and $Y_{i}$ be the success rate of $\pi_{0\rightarrow 1}$ and $\pi_{1}$ in the $i^{th}$ testing environment and define $D_{i} = |Y_{i} - X_{i}|$. 

The null hypothesis $H_{0}$ is established as: there is no significant performance discrepancy between $\pi_{0\rightarrow 1}$ and $\pi_{1}$. $H_{0}$ and the alternative hypothesis $H_{1}$ then write:
\begin{equation}
    H_{0}: \mu_{D}  \leqslant \delta \quad \text{versus} \quad H_{1}: \mu_{D} > \delta
\end{equation}
where $\mu_{D}$ is the mean value of $D_{i}$.

The unbiased standard deviation $\hat{\sigma}_{D}$ is calculated as:
\begin{equation}
    \hat{\sigma}_{D}^{2} = \frac{1}{N-1}\sum_{i=1}^{N} (D_{i}-\hat{\mu}_{D})^{2}
\end{equation}
where $N$ is the number of testing rollouts, $\hat{\mu}_{D}$ is the estimated mean value of $D_{i}$.

According to dependent $t$-test, the $t$-statistic is given by:
\begin{equation}
    t = \frac{\hat{\mu}_{D}-\delta}{\hat{\sigma}_{D}/\sqrt{N}}
\end{equation}

We reject $H_{0}$ if $t$ is greater than the critical value given a preset significance level $\alpha\in (0, 1)$:
\begin{equation}
    t > t_{\alpha, N-1}
\end{equation}

Therefore, if the $t$ value derived through experiments is larger than the critical value, the null hypothesis can be rejected, providing quantitative evidence that visual encoders play an active role in decision-making.
The framework of VAT is demonstrated in \textit{Fig.}~\ref{fig:vat_pipeline}.

\subsection{Full-Gradient Saliency Map}
In addition to the quantitative results of VAT, we adopt \textit{FullGrad}~\cite{Srinivas2019FullGradientRF} to qualitatively demonstrate how visual encoders capture specific task information and contribute to the decision-making process. 
\textit{FullGrad} aggregates full-gradient components to generate saliency maps that highlight the most influential regions in the input pixels affecting the network's output, making it a widely used tool in model explainability within computer vision.
Examples can be seen in \textit{Fig.}~\ref{fig:saliency}.
\begin{table*}[t]
    \vspace{0.3cm}
    \input{tables/exp_out_domain}
    \caption{\textbf{Benchmarking Results.} 
    For a clear comparison, we report the mean and standard deviation of success rates averaged across models with different regression heads (Left) and with different pretrained or E2E-trained encoders (Right). 
    The first and second highest scores are indicated in bold and underlined, respectively.}
    \label{tab:benchmark}
    \vspace{-0.3cm}
\end{table*}

%% file: tables/pretrain_overview.tex
\small
\centering
\scalebox{0.84}{
\begin{tabular}{ccccc}
    \toprule
    \text{Method} & \text{Objective} & \text{Model} & \text{N-o-P (M)} & \text{Dataset}\\
    \midrule
    ImageNet & Cross Entropy & ResNet18 & 11 & ImageNet\\
    R3M & Temporal Contrastive & ResNet18 & 11 & Ego4D\\
    MVP & MAE & ViT-S & 22 & Multi\\
    VIP & Value Optimization & ResNet50 & 23 & Ego4D\\
    \bottomrule
\end{tabular}
}

%% file: tables/task_overview.tex
\centering
\small
\scalebox{0.88}{
    \begin{tabular}{lccccccc}
        \toprule
        \text{Task} & \text{ActD} & \text{ObsD} & \text{Data} & \text{Step} & \text{RichC} & \text{LongH}\\
        \toprule
        \multicolumn{7}{c}{Simulation Benchmark}\\
        \midrule
        \textcolor{blue}{\textit{Stack}} & 7 & 84 & 200 & 400 & No & No\\
        \textcolor{blue}{\textit{Can}} & 7 & 84 & 200 & 400 & No & No\\
        $\textit{Can}^{*}$ & 7 & 84 & 200 & 400 & No & No\\
        Pick Place & 7 & 84 & 200 & 400 & No & Yes\\
        \textcolor{blue}{\textit{Square}} & 7 & 84 & 200 & 400 & Yes & No\\
        $\textit{Square}^{*}$ & 7 & 84 & 200 & 400 & Yes & No\\
        \textit{Nut Assembly} & 7 & 84 & 200 & 400 & Yes & Yes\\
        \midrule
        \textcolor{blue}{\textit{PushT}} & 2 & 96 & 200 & 300 & Yes & Yes\\
        \midrule
        \textit{Mug} & 6 & 224 & 1000 & -- & No & No\\
        \textit{Spam} & 6 & 224 & 1000 & -- & No & No\\
        \toprule
        \multicolumn{7}{c}{Real-world Benchmark}\\
        \midrule
        \textit{Real Reach} & 7 & 84 & 160 & -- & No & No\\
        \textit{Real PnP} & 7 & 84 & 160 & -- & No & No\\
        \bottomrule
    \end{tabular}
}

%% file: pics/transfer_task.tex
\minipage{0.22\textwidth}
    \includegraphics[width=\linewidth]{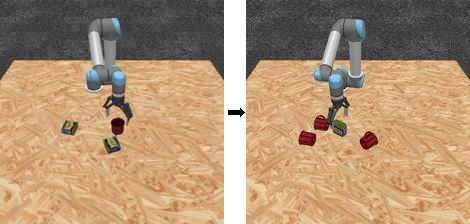}
    \subcaption{\textit{Mug} $\rightarrow$ \textit{Spam}}\label{fig:task_vat_1}
\endminipage\hspace{0.5cm}
\minipage{0.22\textwidth}
    \includegraphics[width=\linewidth]{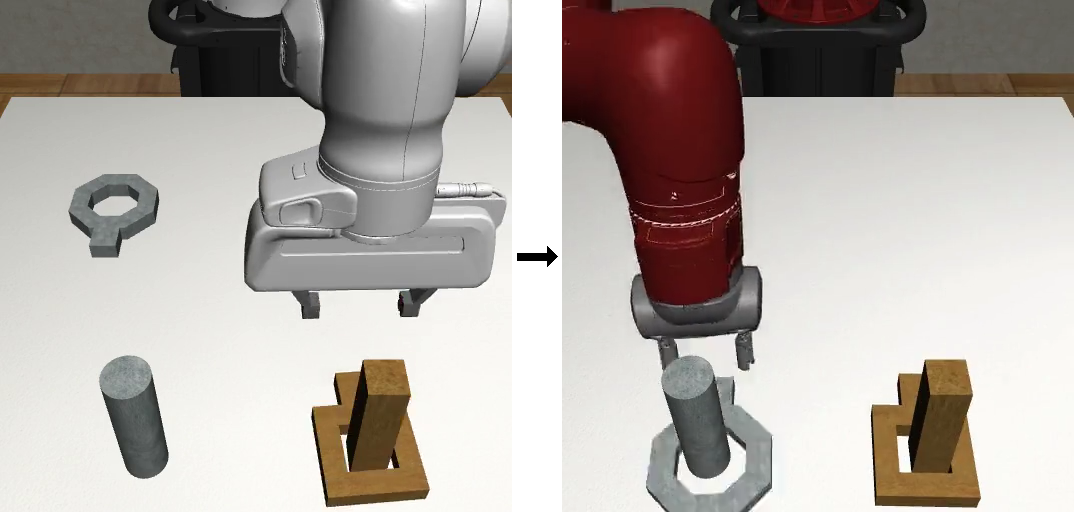}
    \subcaption{$\textit{Square}^{\star}\rightarrow$ \textit{Nut}}\label{fig:task_vat_2}
\endminipage\hfill
\minipage{0.22\textwidth}%
    \includegraphics[width=\linewidth]{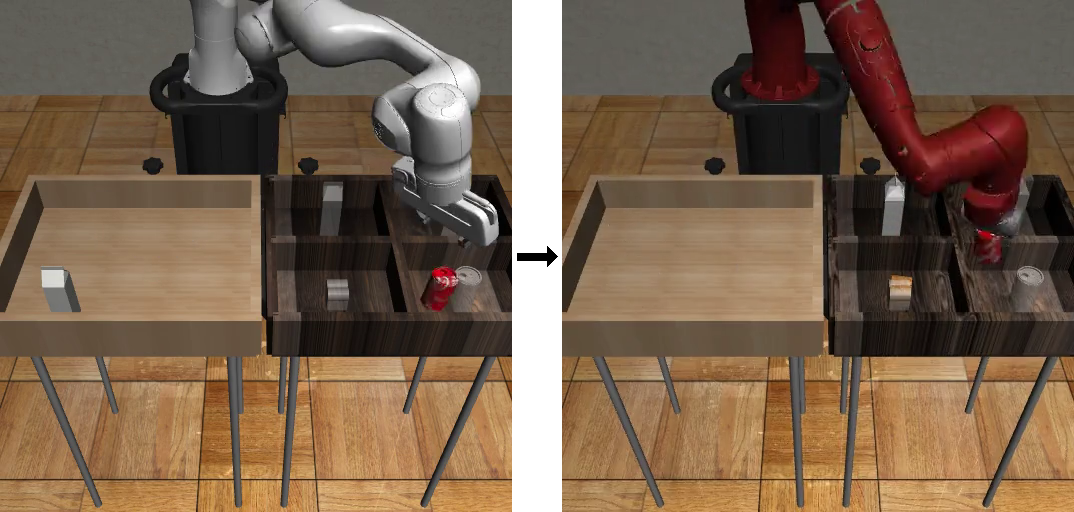}
    \subcaption{$\textit{Can}^{\star}\rightarrow$ \textit{Pick Place}}\label{fig:task_vat_3}
\endminipage\hspace{0.5cm}
\minipage{0.22\textwidth}%
    \includegraphics[width=\linewidth]{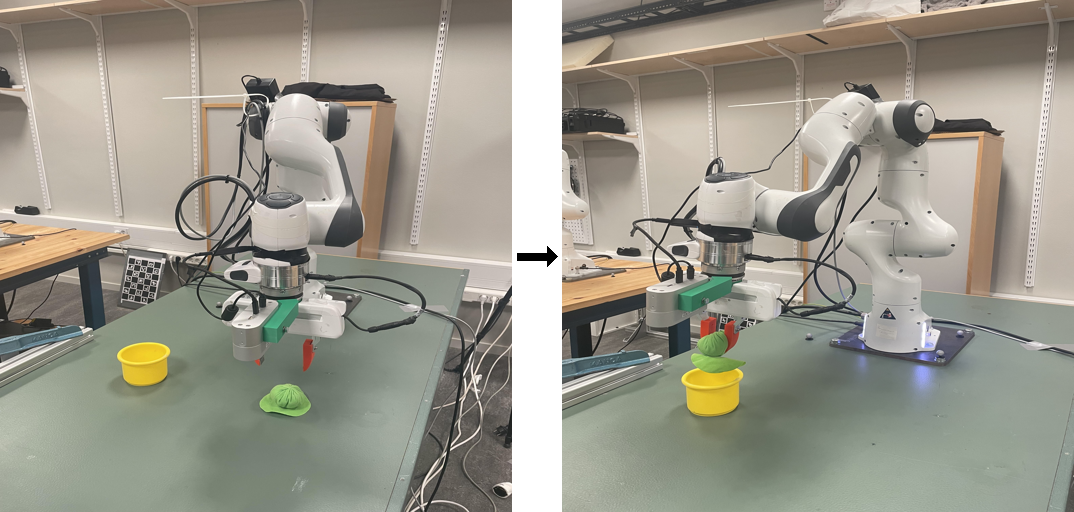}    \subcaption{\textit{Reach} $\rightarrow$ \textit{PnP}}\label{fig:task_vat_4}
\endminipage

%% file: tables/exp_out_domain.tex
\centering
\small
\scalebox{0.98}{
\begin{tabular}{lccccccccc}
     \toprule
     \multicolumn{1}{c}{\textbf{Task}} & \multicolumn{5}{c}{\textbf{Avg. Policy}}& \multicolumn{4}{c}{\textbf{Avg. Encoder}}\\
     \cmidrule(lr){2-6}\cmidrule(lr){7-10}
     & \text{ImageNet} & \text{R3M} & \text{MVP} & \text{VIP} & \text{E2E} & \text{BC} & \text{BC-RNN} & \text{DP-C} & \text{DP-T} \\
     \toprule
     \textit{PushT} & $0.42_{\pm 0.01}$ & $0.51_{\pm 0.01}$ & $\underline{0.54}_{\pm 0.02}$ & $0.19_{\pm 0.01}$ & $\textbf{0.80}_{\pm 0.01}$ & $0.25_{\pm 0.01}$ & $0.45_{\pm 0.01}$ & $\textbf{0.64}_{\pm 0.01}$ & $\underline{0.62}_{\pm 0.01}$\\
     \textit{Can} & $\underline{0.92}_{\pm 0.02}$ & $0.81_{\pm 0.03}$ & $0.72_{\pm 0.02}$ & $0.29_{\pm 0.02}$ & $\textbf{0.96}_{\pm 0.01}$ & $0.55_{\pm 0.03}$ & $0.67_{\pm 0.01}$ & $\underline{0.86}_{\pm 0.02}$ & $\textbf{0.87}_{\pm 0.02}$\\
     \textit{Stack} & $\underline{0.63}_{\pm 0.02}$ & $0.61_{\pm 0.03}$ & $0.47_{\pm 0.03}$ & $0.05_{\pm 0.03}$ & $\textbf{0.91}_{\pm 0.02}$ & $0.34_{\pm 0.03}$ & $0.53_{\pm 0.01}$ & $\underline{0.62}_{\pm 0.03}$ & $\textbf{0.64}_{\pm 0.02}$\\
     \textit{Square} & $\underline{0.59}_{\pm 0.03}$ & $0.33_{\pm 0.03}$ & $0.38_{\pm 0.03}$ & $0.05_{\pm 0.02}$ & $\textbf{0.75}_{\pm 0.02}$ & $0.22_{\pm 0.02}$ & $0.36_{\pm 0.03}$ & $\textbf{0.56}_{\pm 0.03}$ & $\underline{0.53}_{\pm 0.02}$\\
     \bottomrule
\end{tabular}
}

%% file: body/exp.tex
\subsection{Experimental Setup}
\subsubsection{Evaluation Protocol} The evaluation metrics include success rate for all \textit{Robomimic} and \textit{MimicGen} tasks and the real-world experiments, target area coverage~\cite{florence2021implicit} for \textit{PushT} and $L_{1}$ validation loss for \textit{UR5 Reach} tasks.
In simulations, the results are reported as the average of the top 3 best-performing checkpoints saved every 30 epochs. 
Each checkpoint is tested in 50 environment rollouts with different initializations (unseen in the training data), resulting in 150 trials in total.
For real-world experiments, we evaluate the success rate based on 20 robot executions.

\subsubsection{Implementation Detail} In all experiments, images are normalized to $[0, 1]$. Random cropping is uniformly applied with a cropping size of 76$\times$76 pixels, except for the \textit{UR5 Reach} tasks. 
Following the original setups of the OOD-pretrained encoders, we resize images to 224$\times$224 pixels after cropping for all encoders and multiply the images by 255.0 for R3M and VIP. 
Policies are optimized by the Adam optimizer~\cite{adam2015} with a learning rate of $1\times 10^{-4}$ for 600 epochs and a batch size of 128. 
The frozen encoders used in VAT are selected based on the best-performing checkpoint in simulation and the last checkpoint in the real world.

\subsection{Experimental Result}

\noindent\textbf{[Result 1] OOD pretraining formulation does not lead to consistent performance gain.}

We establish a performance benchmark that evaluates four frozen encoders (Imagenet, R3M, MVP and VIP) in combination with four regression heads (BC, BC-RNN, DP-C and DP-T) on four challenging manipulation tasks (\textit{Stack}, \textit{Can}, \textit{Square} and \textit{PushT}). 
We compare these combinations with their corresponding policies that utilize a ResNet-18 as the visual encoder trained end-to-end. 
The results are summarized in \textit{Tab.}~\ref{tab:benchmark}.
As shown in the left part of \textit{Tab.}~\ref{tab:benchmark}, the OOD-pretrained encoders fail to maintain the reported performance margin over E2E-trained models, resulting in an average performance gap of 42\% across all settings. 
This gap exaggerates with increasing task complexity, from simple pick-and-place tasks like \textit{Can} to long-horizon contact-rich tasks, such as \textit{PushT} and \textit{Square}.
Additionally, the right part of \textit{Tab.}~\ref{tab:benchmark} shows that downstream policies with greater capacity significantly enhance model performance for both E2E and pretrained settings.

\begin{table}[ht]
    \centering
    \input{tables/exp_transfer/transfer}
    \caption{\textbf{Results for Visual Alignment Testing.}
    We report the mean and standard deviation of success rates of $\pi_{1}$ and $\pi_{0\rightarrow 1}$ for \textit{Nut Assembly} and \textit{Pick Place} in simulations, as well as \textit{Real PnP} in the real world. 
    VAT is performed on these tasks and the $t$-statistic is compared against critical values of 1.66 and 1.73 in simulation and the real world, respectively, at a significance level of $\alpha = 0.05$. $\delta$ is set to 0.05 based on the standard deviation. The null hypothesis is consistently rejected in all cases.}
    \label{tab:vat_result}
    \vspace{-0.2cm}
\end{table}
\begin{figure}[ht]
    \centering
    \includegraphics[width=0.62\linewidth]{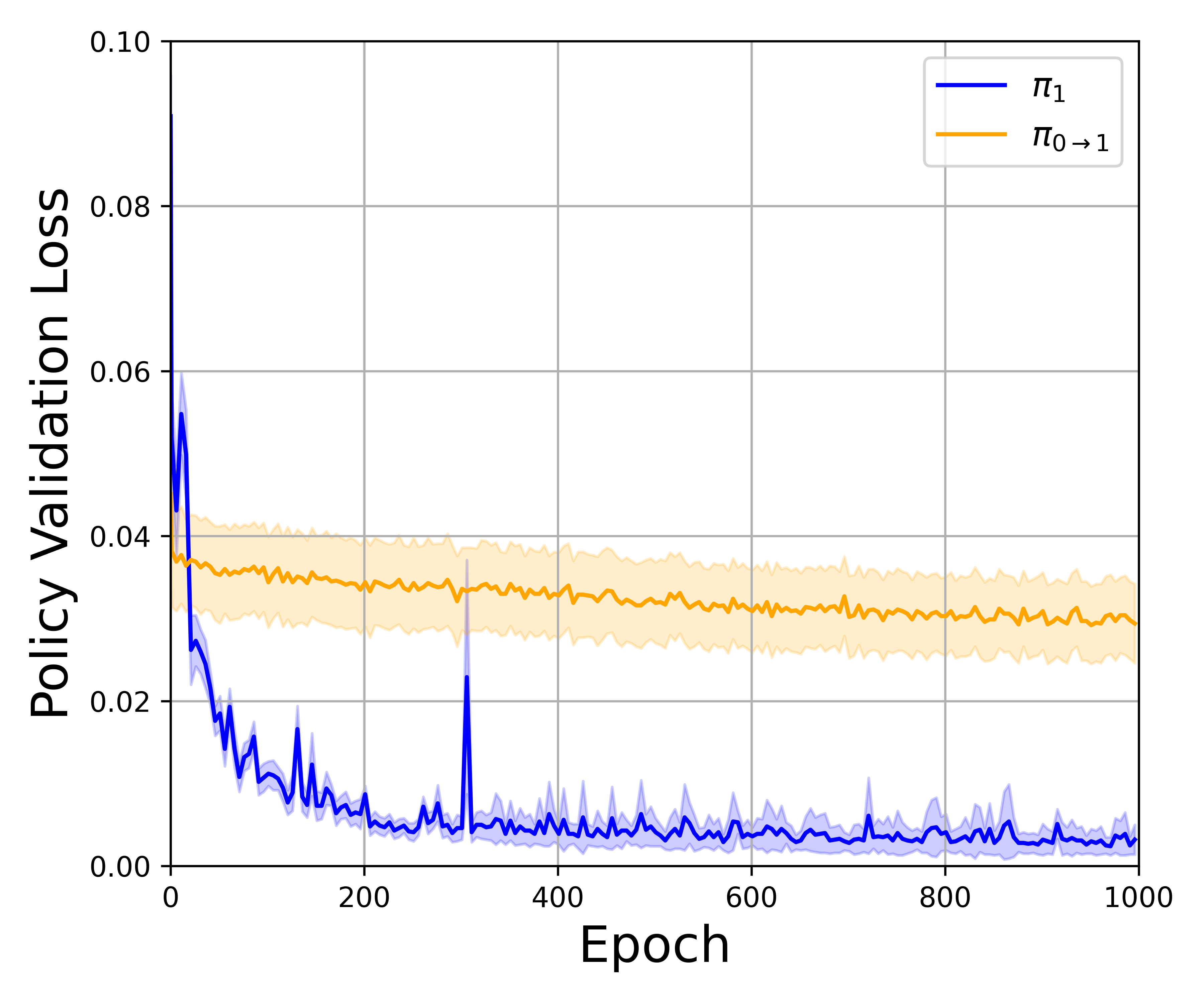}
    \caption{\textbf{Validation Loss of \textit{Spam}.} 
    For reaching tasks, the difference between validation loss of $\pi_{1}$ and $\pi_{0\rightarrow 1}$ indicates their encoders are not interchangeable.}
    \label{fig:vat_loss}
    \vspace{-0.5cm}
\end{figure}
\begin{figure*}[th]
    \centering
    \input{pics/saliency}
    \caption{\textbf{Saliency Maps.}
    We present saliency maps of the E2E-trained and OOD-pretrained encoders in (a); encoders $f_{1}$ and $f_{0}$ of tasks for Visual Alignment Testing in (b); the encoder from an E2E multi-task policy with a shared encoder and three separate regression heads, for the three tasks it was trained on in (c). 
    The saliency map does not apply to the non-CNN-based MVP and it appears inconsistent for VIP, likely because VIP was trained to generate RL rewards. 
    In \textit{Real PnP}, $f_{1}$ only activates on the bucket after the robot picks up the ball, while $f_{0}$ distributes its attention across the entire scene.}
    \label{fig:saliency}
    \vspace{-0.3cm}
\end{figure*}
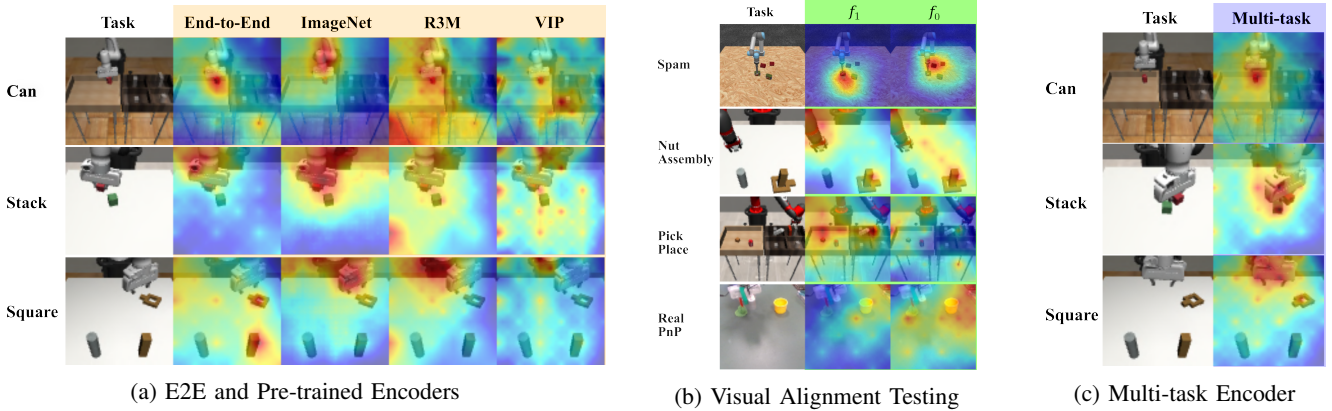

\noindent\textbf{[Result 2] The performance gap stems from E2E-trained visual encoders actively influencing decision-making.}

OOD pretraining differs from E2E training due to its modular formulation of visuomotor policies and the functional separation of visual encoders. The validity of this separation assumes that visual encoders function solely as general feature extractors, without contributing to the decision-making process. However, the benchmarking results suggest that this assumption may not fully hold in practice.

\noindent\textit{Quantitative Result.} To further explore the role of visual encoders, we perform Visual Alignment Testing on four groups of tasks, as illustrated in \textit{Fig.}~\ref{fig:vat_task}. 
With a significance level at $\alpha = 0.05$, the null hypothesis is consistently rejected in \textit{Nut Assembly}, \textit{Pick Place} and \textit{Real PnP} when using both BC and DP-C as regression heads, as is summarized in \textit{Tab.}~\ref{tab:vat_result}. 
For the \textit{UR5 Reach} task, the significant difference in validation loss between $\pi_{1}$ and $\pi_{0\rightarrow 1}$ in \textit{Fig.}~\ref{fig:vat_loss}, demonstrates that the encoders are not interchangeable.
Following the arguments in section III.C, these results provide quantitative evidence that, E2E-trained visual encoders play an active role in decision-making in the overall policy. 
Additionally, the performance gap between $\pi_{1}$ and $\pi_{0\rightarrow 1}$ as well as between E2E and OOD-pretrained models in the benchmark, indicates that this decision-making ability of visual encoders, acquired naturally in E2E training through specific task motor data supervision, positively impacts overall policy performance.

\noindent\textit{Qualitative Result.} To understand how visual encoders contribute to decision-making, we present the saliency maps of encoders $f_{1}$ and $f_{0}$ trained on VAT tasks in \textit{Fig.}~\ref{fig:saliency_2}. 
Notably, \textit{E2E-trained encoders are primarily activated by specific task-relevant areas.}
For example, given the same input image of \textit{Spam}, E2E-trained $f_{1}$ mainly focuses on the box of spam the robot is reaching, while $f_{0}$ (trained on \textit{Mug}) focuses on the mugs. 
Despite identical visual inputs, encoders updated with different motor data develop preferences that align with task intent.
Comparing the saliency maps of E2E and OOD-pretrained encoders in \textit{Fig.}~\ref{fig:saliency_1}, the underperforming of pretrained encoders might come from their failure to focus on task-related areas effectively. 
For instance, in \textit{Square}, the E2E model activates on the center of the square nut and the rod, whereas Imagenet, R3M and VIP do not give sufficient attention to the task-relevant objects. 

\begin{table}[ht]
    \vspace{-0.2cm}
    \centering
    \scalebox{0.9}{
    \begin{tabular}{cccc}
        \toprule
         \textbf{Encoder} & \text{E2E} & \text{Pretrain} & \text{Pretrain+Finetune}\\
         \toprule
         \textbf{Succ. rate} & 0.68 & 0.26 & 0.61\\
         \bottomrule
    \end{tabular}
    }
    \caption{\textbf{Finetuning Results.} All finetuned models are trained and evaluated following the same protocol in Section IV.A. All models use BC as the regression head and averaged over $\textit{PushT}$, $\textit{Can}$ and $\textit{Square}$. The results of pretrained encoders are averaged over ImageNet, R3M and VIP.}
    \label{tab:finetune}
    \vspace{-0.3cm}
\end{table}

In \textit{Fig.}~\ref{fig:saliency_3}, we demonstrate that the decision-making ability of E2E encoders extends to multi-task settings, where they differentiate between tasks and focus on task-relevant features accordingly.
This alleviates concerns that their decision-making ability results from overfitting to a single task.
Furthermore, \textit{Fig.}~\ref{fig:saliency_1} shows that ImageNet features effectively recognize large objects in a scene, capturing most key information. This non-specific yet information-rich property likely explains ImageNet's superior performance compared to other pretrained encoders.
For completeness, \textit{Tab.}~\ref{tab:finetune} shows that finetuning pretrained encoders can enhance their performance, which does not contradict our conclusions. Investigating optimal finetuning methods and how they influence the representations lies beyond the scope of this study.

%% file: tables/exp_transfer/transfer.tex
\small
\scalebox{0.9}{
\begin{tabular}{l|ccccc}
    \toprule
    \text{Task} &\text{Policy} & $\pi_{0\rightarrow 1}$ & $\pi_{1}$ & $t$ & \text{Rej.} $H_{0}$\\
    \toprule
    & \text{BC} & $0.05_{\pm 0.01}$ & $0.35_{\pm 0.06}$ & 7.61 & \text{Yes}\\
    \multirow{-2}{*}{\textit{Nut Ass.}} & \text{DP-C} & $0.35_{\pm 0.04}$ & $0.54_{\pm 0.01}$ & 10.51 & \text{Yes}\\
    \midrule
    & \text{BC} & $0.01_{\pm 0.00}$ & $0.12_{\pm 0.01}$ & 3.17 & \text{Yes}\\
    \multirow{-2}{*}{\textit{Pick Place}}& \text{DP-C} & $0.10_{\pm 0.01}$ & $0.36_{\pm 0.01}$ & 7.20 & \text{Yes}\\
    \midrule
    &\text{BC} & $0.00_{\pm 0.00}$ & $0.70_{\pm 0.03}$ & 8.17 & \text{Yes}\\
    \multirow{-2}{*}{\textit{Real PnP}}& \text{DP-C} & $0.00_{\pm 0.00}$ & $1.00_{\pm 0.00}$ & $\infty$ & \text{Yes}\\
    \bottomrule
\end{tabular}
}

%% file: pics/saliency.tex
\centering
\minipage{0.42\linewidth}
    \scalebox{1.0}{
        \includegraphics[width=\linewidth]{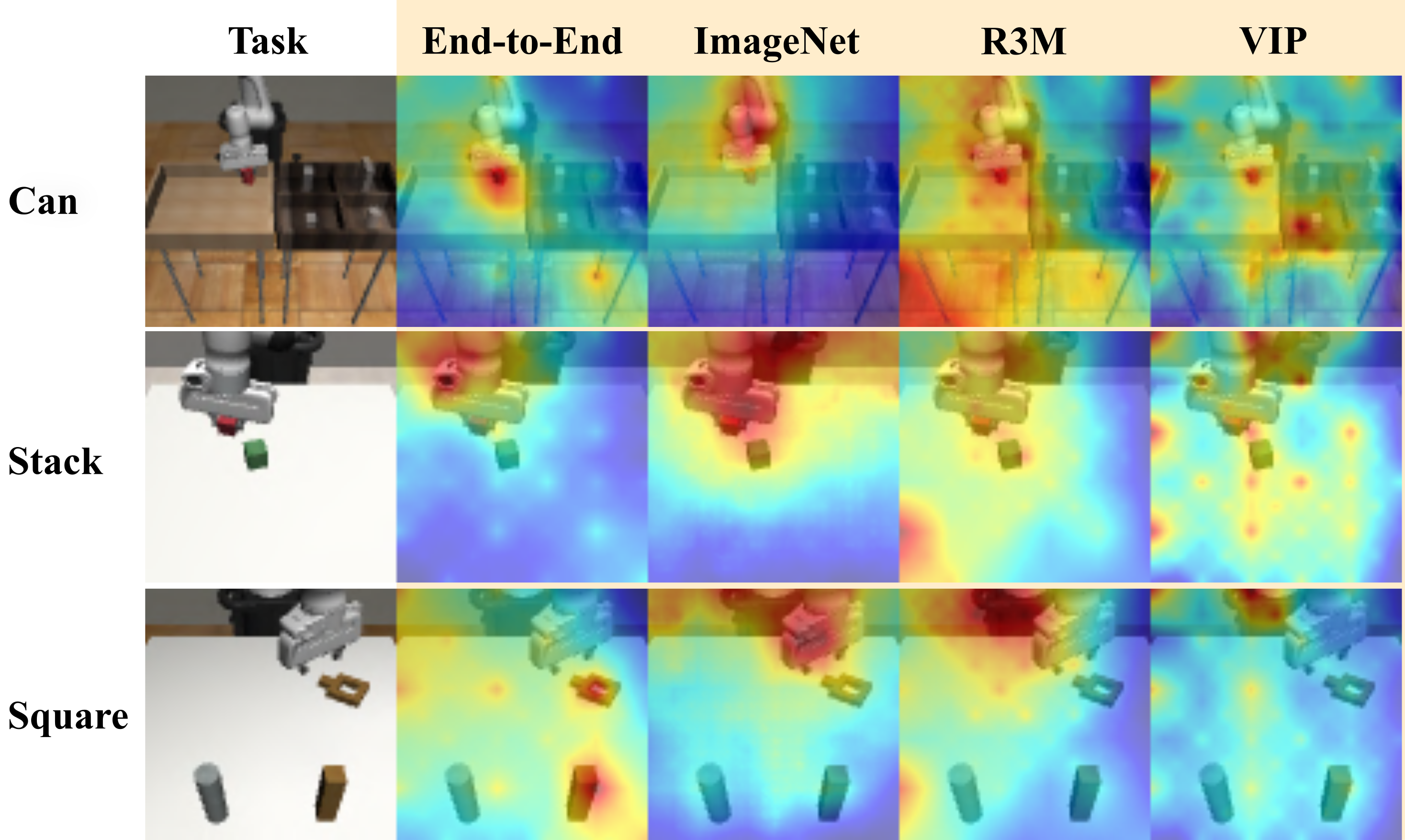}
    }
    \subcaption{E2E and Pretrained Encoders}
    \label{fig:saliency_1}
\endminipage\hspace{0.5cm}
\minipage{0.22\linewidth}
    \scalebox{0.95}{
        \includegraphics[width=\linewidth]{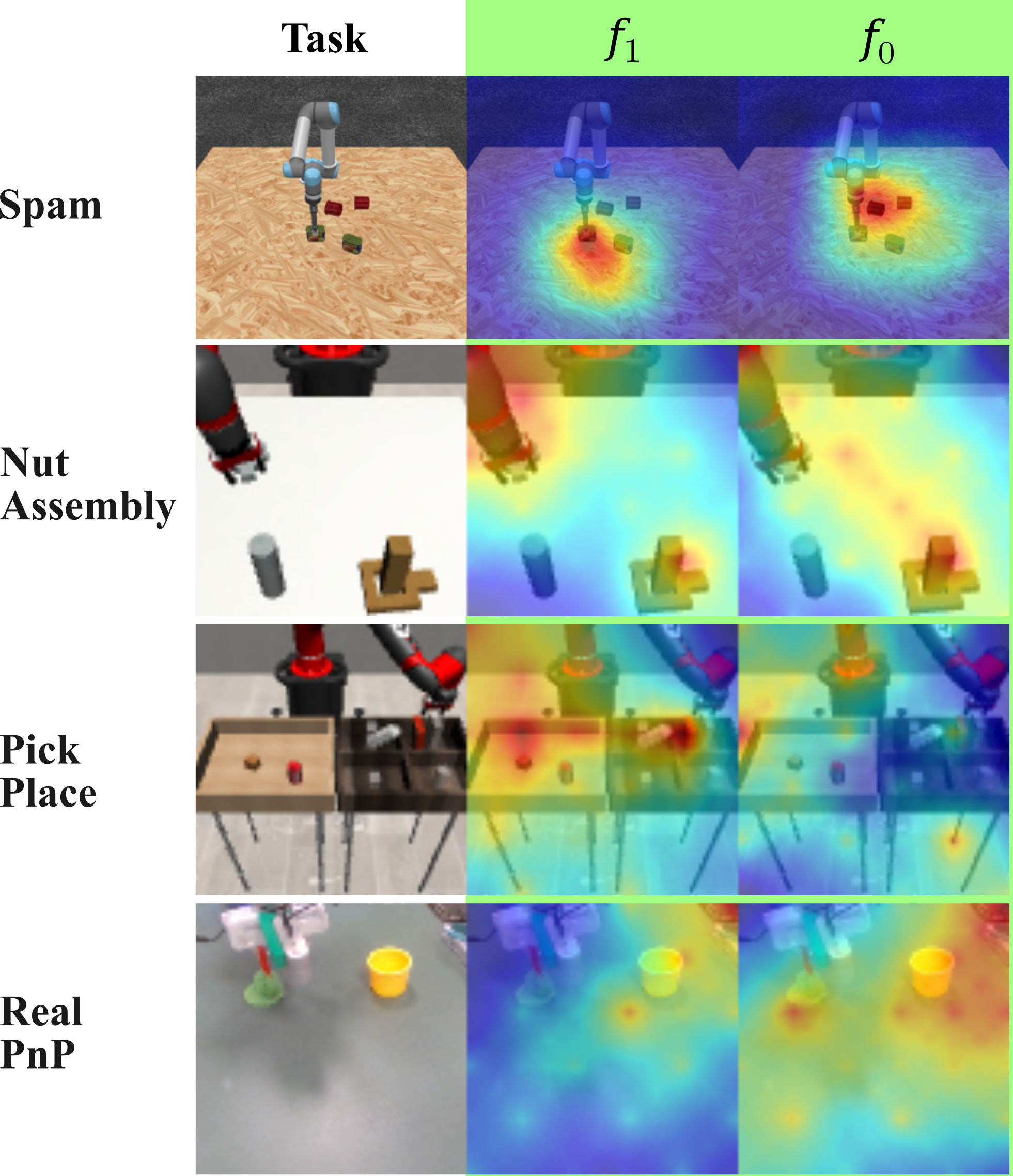}
    }
    \subcaption{Visual Alignment Testing}
    \label{fig:saliency_2}
\endminipage\hspace{0.5cm}
\minipage{0.2\linewidth}%
    \scalebox{0.95}{
        \includegraphics[width=\linewidth]{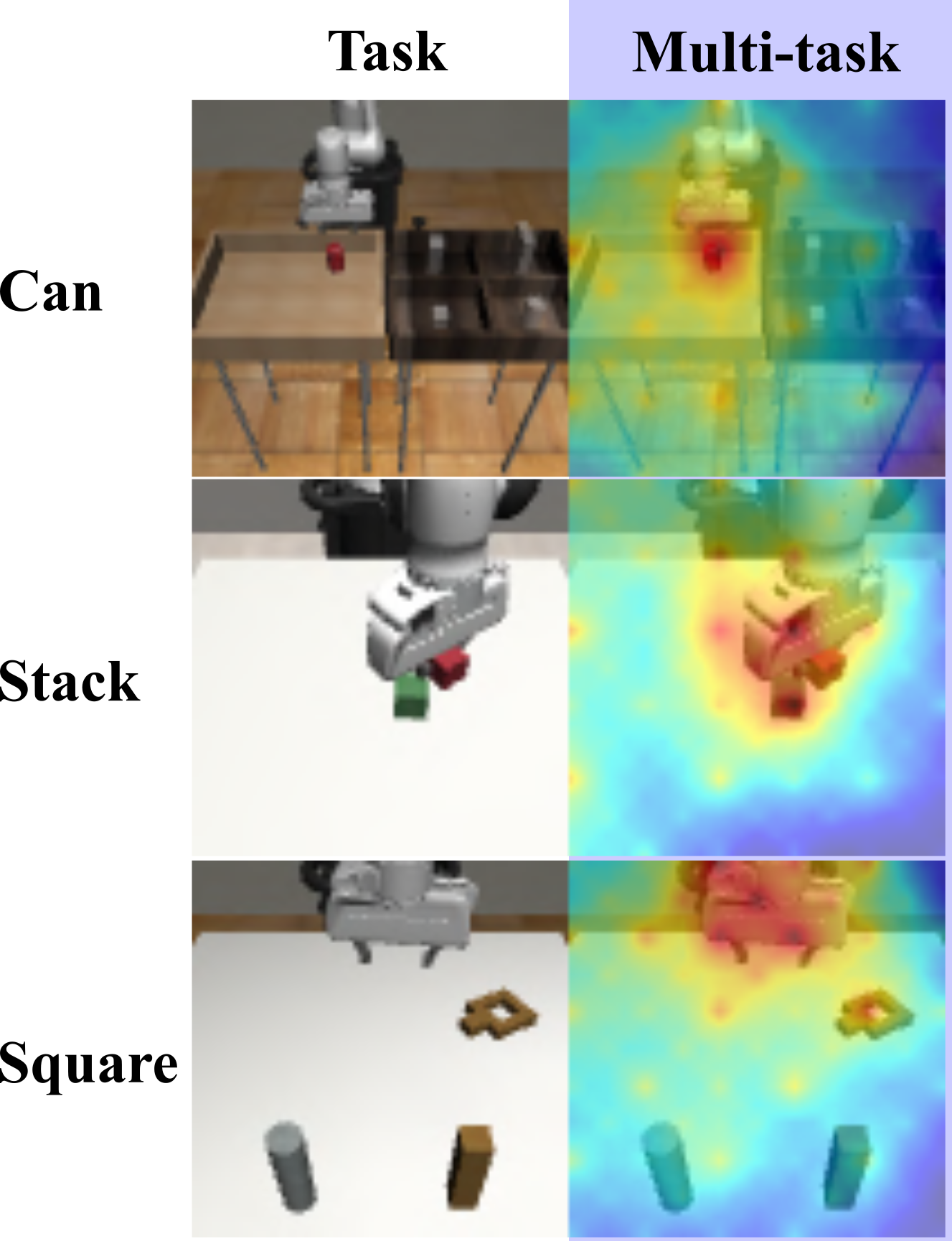}
    }
    \subcaption{A Multi-task Encoder}
    \label{fig:saliency_3}
\endminipage

%% file: body/conclusion.tex
In this paper, we identify and analyze an underexplored question arising from the introduction of OOD pretraining to visuomotor policy learning: \textit{E2E-trained visual encoders actively contribute to the decision-making process by focusing on specific task-relevant areas in the visual inputs.} Both quantitative evidence derived from the proposed Visual Alignment Testing, applied in simulations and the real world, and qualitative results provided by \textit{FullGrad} saliency maps support this claim.
This finding provides an initial explanation for the underperformance of OOD pretrained encoders, as demonstrated in the benchmarking results.
Our results indicate that future works may benefit from training visual encoders on manipulation datasets that span diverse task domains and incorporating task conditions and context information into the encoders.

One limitation of this work is the exclusion of very low-data regimes explored in some pretraining studies, e.g. working with 5 demonstrations. 
A significant reduction in training data is likely to impair the decision-making ability of the E2E visual encoders. Investigating the effects of smaller dataset sizes and identifying the threshold at which performance degradation becomes significant is an interesting avenue for future research.
Additionally, our analysis is confined to the encoder level. 
Examining how the policy further leverages these representations is another direction worth exploring.